\documentclass[sn-mathphys-num]{sn-jnl}

\usepackage{graphicx}%
\usepackage{multirow}%
\usepackage{amsmath,amssymb,amsfonts}%
\usepackage{amsthm}%
\usepackage{mathrsfs}%
\usepackage[title]{appendix}%
\usepackage{xcolor}%
\usepackage{textcomp}%
\usepackage{manyfoot}%
\usepackage{booktabs}%
\usepackage{algorithm}%
\usepackage{algorithmicx}%
\usepackage{algpseudocode}%
\usepackage{listings}%

\theoremstyle{thmstyleone}%
\theoremstyle{thmstyletwo}%

\theoremstyle{thmstylethree}%

\begin{document}

\title[Article Title]{Adaptive Path-Planning for Autonomous Robots: A UCH-Enhanced Q-Learning Approach}


\author*[1,2,3]{\fnm{Wei} \sur{Liu}}\email{liuwei@lntu.edu.cn}

\author[1]{\fnm{Ruiyang} \sur{Wang}}\email{472321492@stu.lntu.edu.cn}
\equalcont{These authors contributed equally to this work.}

\author[4]{\fnm{Haonan} \sur{Wang}}\email{hwang298@jh.edu}
\equalcont{These authors contributed equally to this work.}

\author[5]{\fnm{Guangwei} \sur{Liu}}\email{liuguangwei@lntu.edu.cn}
\equalcont{These authors contributed equally to this work.}

\affil[1]{\orgdiv{College of Science}, \orgname{Liaoning Technical University}, \orgaddress{ \city{Fuxin}, \postcode{123000}, \state{Liaoning}, \country{China}}}

\affil[2]{\orgdiv{Institute of Mathematics and Systems Science}, \orgname{Liaoning Technical University}, \orgaddress{ \city{Fuxin}, \postcode{123000}, \state{Liaoning}, \country{China}}}

\affil[3]{\orgdiv{Institute of Intelligent Engineering and Mathematics}, \orgname{Liaoning Technical University}, \orgaddress{ \city{Fuxin}, \postcode{123000}, \state{Liaoning}, \country{China}}}

\affil[4]{\orgdiv{Whiting School of Engineering}, \orgname{Johns Hopkins University}, \orgaddress{\city{Baltimore}, \postcode{21218}, \state{Maryland}, \country{USA}}}

\affil[5]{\orgdiv{College of Mines}, \orgname{Liaoning Technical University}, \orgaddress{ \city{Fuxin}, \postcode{123000}, \state{Liaoning}, \country{China}}}


\abstract{
Q-learning methods are widely used in robot path planning but often face challenges of inefficient search and slow convergence. We propose an Improved Q-learning (IQL) framework that enhances standard Q-learning in two significant ways. First, we introduce the Path Adaptive Collaborative Optimization (PACO) algorithm to optimize Q-table initialization, providing better initial estimates and accelerating learning. Second, we incorporate a Utility-Controlled Heuristic (UCH) mechanism with dynamically tuned parameters to optimize the reward function, enhancing the algorithm's accuracy and effectiveness in path-planning tasks. Extensive experiments in three different raster grid environments validate the superior performance of our IQL framework. The results demonstrate that our IQL algorithm outperforms existing methods, including FIQL, PP-QL-based CPP, DFQL, and QMABC algorithms, in terms of path-planning capabilities.}




\maketitle

\section{Introduction}\indent

With the rapid development of the combination of control technology and the Artificial Intelligence(AI) field, the intelligent control of mobile robots and their applications like industrial manufacturing, logistics sorting, etc. in this field is evolving towards self-learning and adaptation \cite{ref1}. For example, intelligent control of mobile robots in complex environments can autonomously move in various environments without external assistance \cite{ref2}, which requires navigation \cite{ref3} and motion planning-related technologies in practical applications. Motion planning is divided into path planning and trajectory planning \cite{ref4}. Path planning often serves as the crucial step of trajectory planning, its goal is to find the optimal path from a starting point to an endpoint in a given environment. However, path planning in dynamic environments is more practical and challenging \cite{ref5}.

In recent years, several path-planning algorithms have been widely adopted in the field of mobile robotics, such as Dijkstra combined with octagonal search to optimize paths \cite{ref6}, Bellman-Ford to cope with fuzzy image environments \cite{ref7}, A-Star based on geometrical optimization \cite{ref8}, RRT combined with CNNs to improve the efficiency \cite{ref9}, the Ant Colony Optimization (ACO)-Artificial Potential Field (APF) fusion algorithm for UUV dynamic path planning \cite{ref10}, Particle Swarm Optimization (PSO) combined with higher-order Bessel curves to achieve smooth paths \cite{ref11} and improved Genetic Algorithm (GA) to cope with complex maps \cite{ref12}. In addition, many researchers also use Reinforcement Learning (RL) to address modeling unknowns and accelerate convergence \cite{ref13}, which presents G2RL for large dynamic environments and introduces MAPPER \cite{ref14}, a decentralized evolutionary RL for hybrid dynamics that combines DNNs and MDPs for UAV autonomy \cite{ref15}. DRL is utilized in DDQN with map data to balance navigation and mission goals \cite{ref16}. SAC is applied with DRL for robotic arm obstacle avoidance \cite{ref17}. Two-depth Q-networks are proposed for QoS-driven actions \cite{ref18}. An optimized TD3 model is offered for UAV energy-efficient paths \cite{ref19}. DQN-based DRL is used to enhance USV autopilots and collision avoidance \cite{ref20}. DDPG and priority sampling are leveraged in a COLREGs-based USV avoidance algorithm \cite{ref21}. MADPG and Gumbel-Softmax are employed for AGV conflict-free paths \cite{ref22}. DRL is proposed for real-time planning of driverless vehicles in challenging settings \cite{ref23}.

Although the above algorithms have achieved some results, they still face significant challenges such as difficult environment modeling, poor adaptation to complex environments, slow convergence of algorithms, easy falling into local optimums, high consumption of computational resources, and difficulty in parameter tuning. Meanwhile, researchers have extensively applied Q-learning and its enhancements to path planning in RL, achieving notable success. DFQL \cite{ref24} marries Q-learning with artificial potential fields, effectively tackling UUV path planning in partly known seas. EQL \cite{ref25} expedites convergence to optimal paths for mobile robots via innovative rewards, enhancing path optimization, efficiency, and safety. However, Q-table initialization \cite{ref26} crucially impacts performance, prompting Hao Bing et al. \cite{ref27} to optimize it with FPA for quicker path searches. Heuristic algorithms \cite{ref28,ref29} address complex problems but grapple with non-optimality and complexity, requiring enhancements for efficiency and learning speed. Q-learning faces credit allocation hurdles, local optima traps, dimensionality challenges, and reward function over-optimization \cite{ref30,ref31,ref32,ref33,ref34}, hindering real-time action assessment, strategy discovery, model performance, and generalization. To tackle these, researchers devised refined reward functions \cite{ref35}, inspired by game theory's utility theory \cite{ref36} and hierarchical/model-based RL \cite{ref37}. Meiyan Zhang et al. \cite{ref38} proposed a Predator-Prey-based reward to boost performance. While advancements have been made, mobile robot path planning in dynamic environments still requires deeper research on adaptability and generalization \cite{ref39}.

Therefore, the motivation of this paper hinges on the shortcomings of traditional Q-learning algorithms, including convergence issues, local optima from under-exploration, slower progress from over-exploration, increased computational demands due to state-space dimensionality, and challenges in learning effective strategies from sparse or poorly designed reward functions.

In this work, we propose a framework for an improved Q-learning algorithm with UCH mechanism and optimized Q-table initialization is proposed to enhance path planning efficiency and selection accuracy.

Our \textbf{research contributions} are outlined as follows:

\begin{itemize}  
    \item We developed a novel framework--\textbf{IQL}, which enhances performance by dynamically adjusting pheromone volatility, thus avoiding local optima.
    \item Our framework improves the Q-table initialization process of the Q-learning algorithm using an enhanced ACO algorithm.
    \item Our framework uses a UCH reward mechanism is introduced that dynamically adjusts reward parameters, reducing exploration dilemmas and enabling more precise reward criterion assessment, thereby enhancing algorithm performance.
\end{itemize}

The subsequent sections of this document are structured as follows: Section 2 covers RL basics, Q-learning, and env modeling for path planning. Section 3 presents PACO for Q-table init, UCH for reward tuning, IQL algorithm, and evaluation. Section 4 discusses simulations and results. Section 5 summarizes key contributions.

\begin{figure}[h] 
    \centering 
    \includegraphics[width=0.65\textwidth]{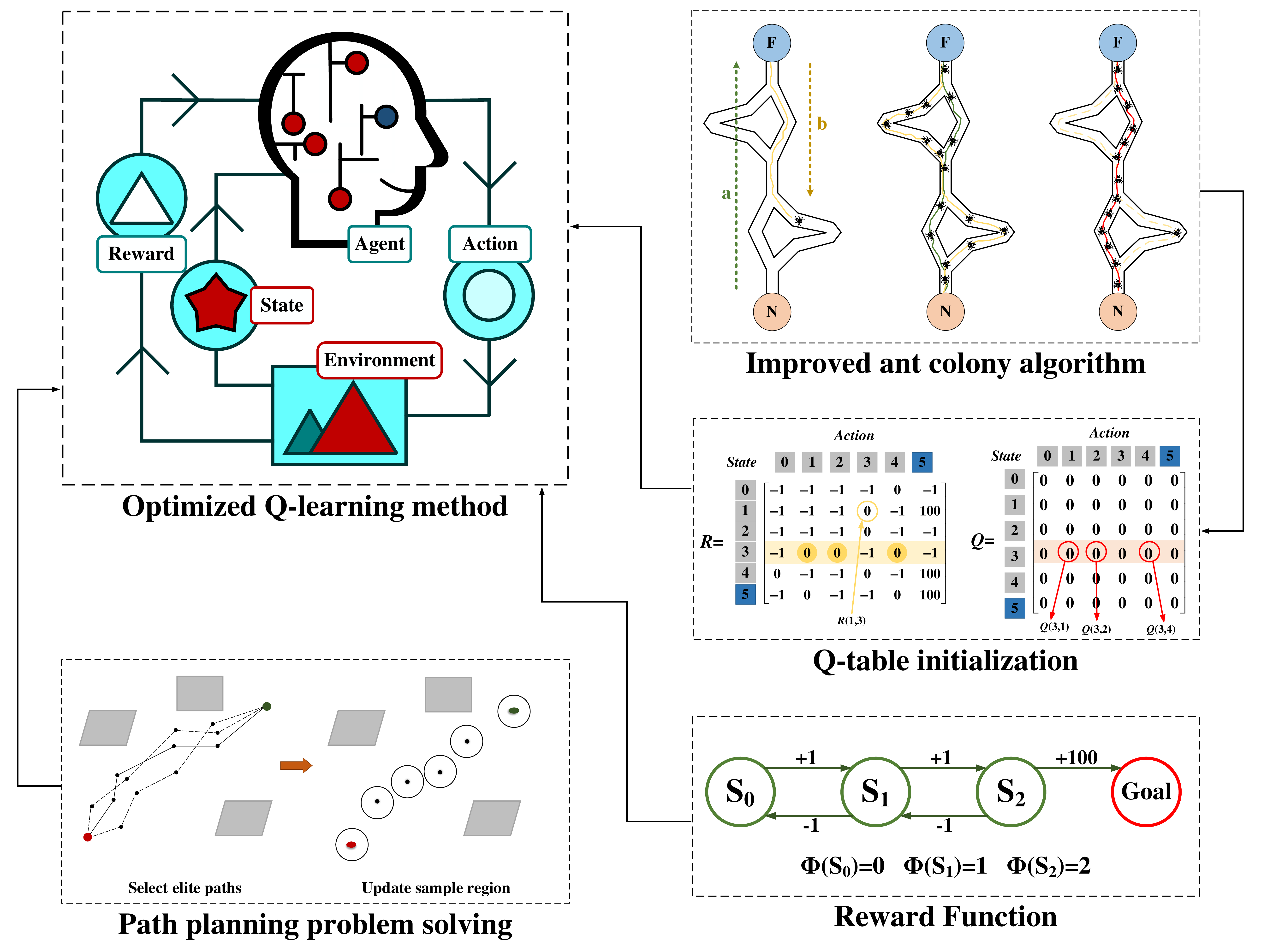} 
    \caption{Framework diagram of the improved Q-learning (IQL) algorithm}
    \label{fig:FIG4}
\end{figure}

\section{Background}\indent

This section covers the basics of RL and traditional Q-learning, focusing on reward functions and environment modeling essential for the improved Q-learning (IQL) algorithm proposed in this paper.

\subsection{Reinforcement Learning}\indent

In the standard RL model, its fundamental framework is composed of three core elements: state, action, and reward. Agent \cite{ref40} acquires the current state \( s _ { t }\) of the environment through observation at moment \textit{t}, and then performs a specific behavior \textit{a} from all feasible behaviors according to an established strategy and receives a reward \( r _ { t }\) for evaluating the merit of the behavior. Subsequently, the intelligent body transfers to a new state \( s _ { t+1 }\) based on the performed behavior and continues the process until the training is complete. The ultimate goal of the intelligent body is to learn a strategy \( S \rightarrow A\) that results in maximizing the desired cumulative reward \textit{r}. In this process, the set of states of the environment is denoted as \(S=\left\{ s _ { 1 }, s _ { 2 }, \ldots, s _ { t + 1 },\ldots \right\}\), while all the possible behaviors constitute the set \(A=\left\{ a _ { 1 }, a _ { 2 }, \ldots, a _ { n } \right\}\). Each behavior has a corresponding Q-value by which the intelligent body decides its behavioral choices.

\subsection{Basic Q-learning algorithms}\indent

Q-learning algorithm \cite{ref41}, introduced in 1989, is a model-free reinforcement learning method that guides an agent's actions across various states. As a classic algorithm, it doesn't require model construction and consists of three main components \cite{ref42}: Q-table initialization, action selection strategy, and Q-table update. Typically, the Q-table is initialized with constant values, and the \( \epsilon \)-greedy strategy is used for action selection. The updated formula for the value function is shown in equation (1).

\begin{equation}
Q(s,a) = Q(s,a) + \alpha \left( {r(s,a) + \gamma \mathop {\max }\limits_{a'} Q(s',a') - Q(s,a)} \right)
\end{equation}

In addition to the three essential components, defining the reward function and collision handling is crucial for the path planning problem. The agent receives a penalty for each step taken, which is related to the length of the path from the previous step. When the agent hits an obstacle, it remains in place and seeks actions to avoid the obstacle until reaching the endpoint. The reward function can be defined as shown in equation (2).

\begin{equation}
 r ( s , a ) = - \sqrt { ( x _ { s } - x _ { s } ^ { \prime \prime } ) ^ { 2 } + ( y _ { s } - y _ { s } ^ { \prime \prime } ) ^ { 2 } }
\end{equation}
where \textit{s} is the current agent state, and  \( x _ { z } , y _ { z } \) is the coordinates of the current state corresponding to the raster; \( s ^ { \prime \prime }\) is the previous state of the agent, and \( ( x _ { z } ^ { \prime \prime }, y _ { z } ^ { \prime \prime } )\) is the coordinates of the raster corresponding to the previous state.

Q-learning is widely used in gaming, robotics, and resource management due to its simplicity and effectiveness, providing a solid foundation for reinforcement learning. However, traditional Q-learning faces challenges such as slow iteration speed, difficulty in environment comprehension during Q-table initialization, and a tendency to fall into local optima.

\begin{figure}[htbp]
\centering
\includegraphics[width=0.7\textwidth]{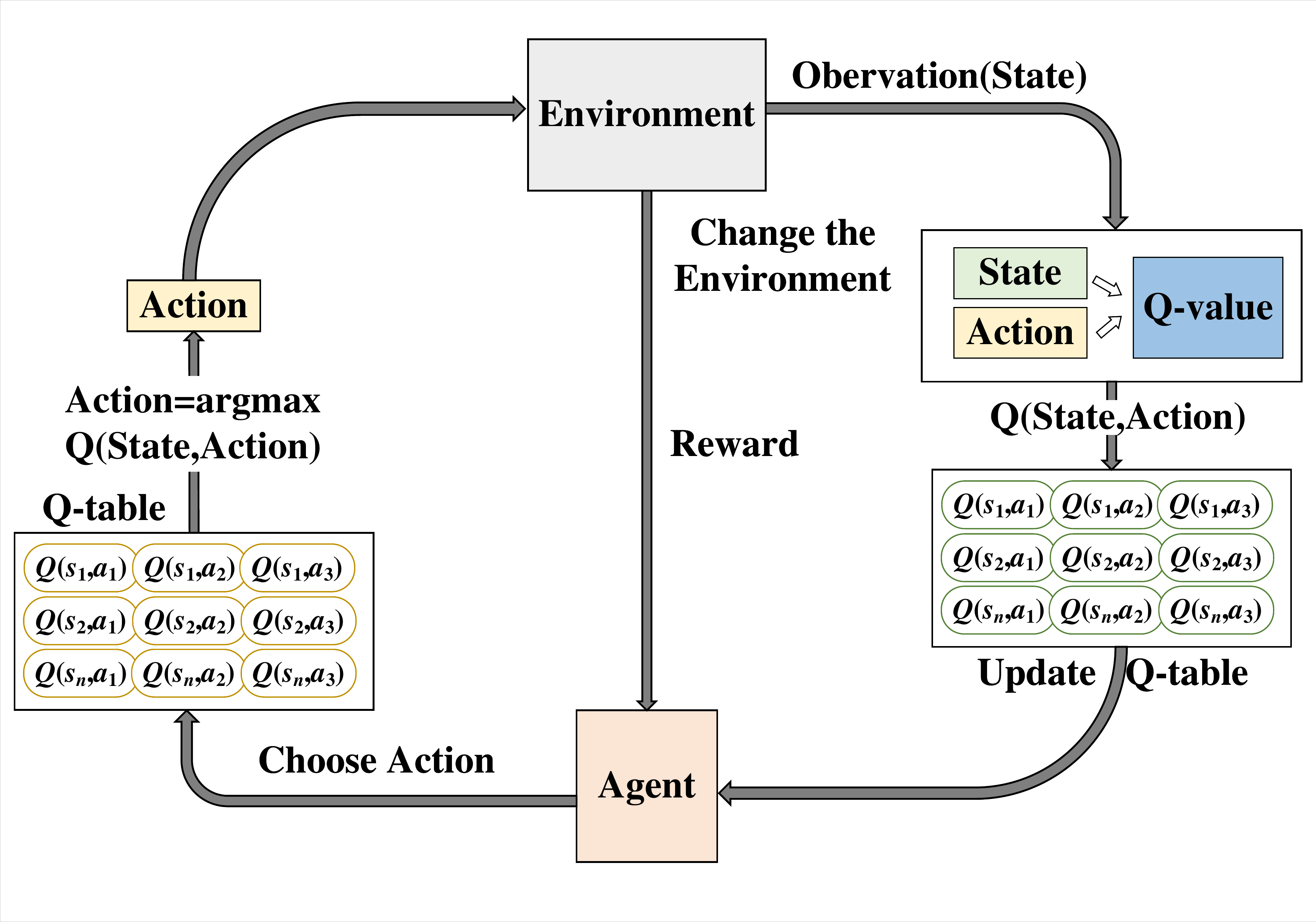}
\caption{Interplay process of the Q-learning algorithm and the surrounding environment}
\label{fig:FIG5}
\end{figure}

The basic flow of the Q-learning algorithm is depicted in \hyperref[fig:FIG5]{Fig. 2}.

\subsection{Environmental modelling for path planning}\indent

This paper uses the raster method to model the environment, which includes static obstacles\cite{ref43}. A grid cell that contains an obstacle is marked as "1" and represented by a black grid in the simulation drawing, regardless of whether the obstacle completely covers the cell. Conversely, a grid cell that has no obstacle is marked as "0" and represented by a white grid in the simulation drawing. 

\begin{figure}[htbp]
\centering
\includegraphics[width=0.45\textwidth]{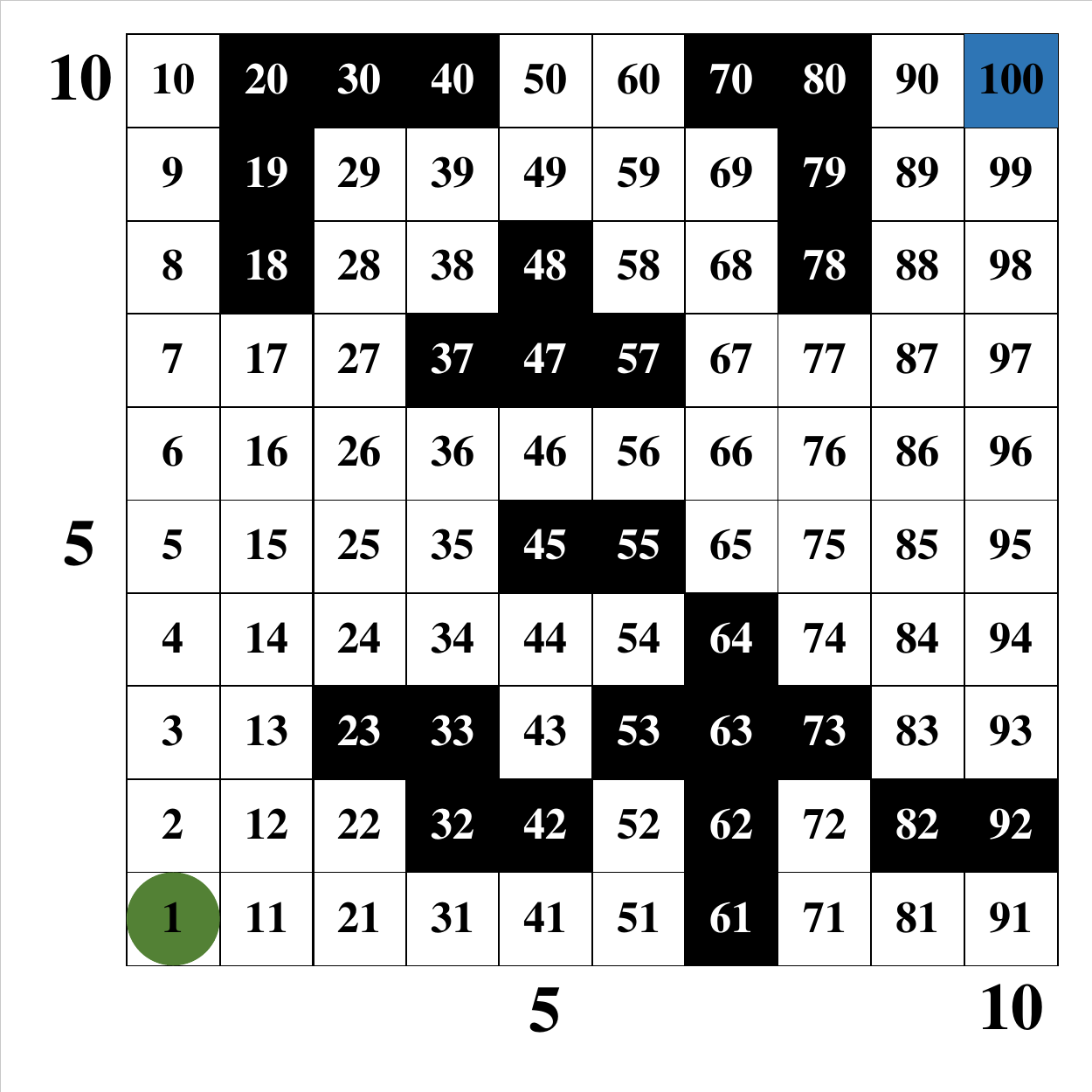}
\caption{Raster map environment and serial number encoding}
\label{fig:FIG1}
\end{figure}

\hyperref[fig:FIG1]{Fig. 3} is used as an example to illustrate the relationship between raster coding and horizontal and vertical coordinates \cite{ref44}. Firstly, a Cartesian coordinate method is used to number grids: horizontal number of grid is used as the abscissa of grid, and vertical number of grid is used as the ordinate of grid. In \hyperref[fig:FIG1]{Fig. 3}, the maximum \textit{h} of the raster abscissa is 10, the maximum \textit{v} of the raster ordinate is 10, and the total number of rasters is 100. Grids are numbered \( 1, 2, \cdots, h \times v\) from bottom to top and left to right. The relationship between the grid coordinates and the grid number is expressed as the following formula. Table 1 shows the symbols and their meanings in (3).

\begin{equation}
\left\{ \begin{array}{l}
{x_c} = ceil({c \mathord{\left/
 {\vphantom {c h}} \right.
 \kern-\nulldelimiterspace} h});\\
{y_c} = \left\{ \begin{array}{l}
\bmod ({c \mathord{\left/
 {\vphantom {c h}} \right.
 \kern-\nulldelimiterspace} h}),\bmod ({c \mathord{\left/
 {\vphantom {c h}} \right.
 \kern-\nulldelimiterspace} h}) \ne 0;\\
h\quad \quad \quad \,\,,\bmod ({c \mathord{\left/
 {\vphantom {c h}} \right.
 \kern-\nulldelimiterspace} h}) = 0.
\end{array} \right.
\end{array} \right.
\end{equation}

\begin{table}[htbp]
  \centering
  \caption{Explanation of the formula symbols for the relationship between grid coordinates and grid numbers}
   \centering
  \begin{tabular}{c c}
    \hline
    Symbols & Symbolic meanings \\
    \hline
    \({x_c}\) & Grid abscissa \\
    \({y_c}\) & Grid ordinates \\
    ceil(\textit{x}) & Take an integer operation that is not less than \textit{x} \\
    mod(\textit{x},\textit{y}) & Take the remainder of \({x \mathord{\left/ {\vphantom {x y}} \right. \kern-\nulldelimiterspace} y}\) \\
    \textit{c} & The current number of the raster \\
    \hline
  \end{tabular}
\end{table}

\begin{figure}[htbp]
\centering
\includegraphics[width=0.36\textwidth]{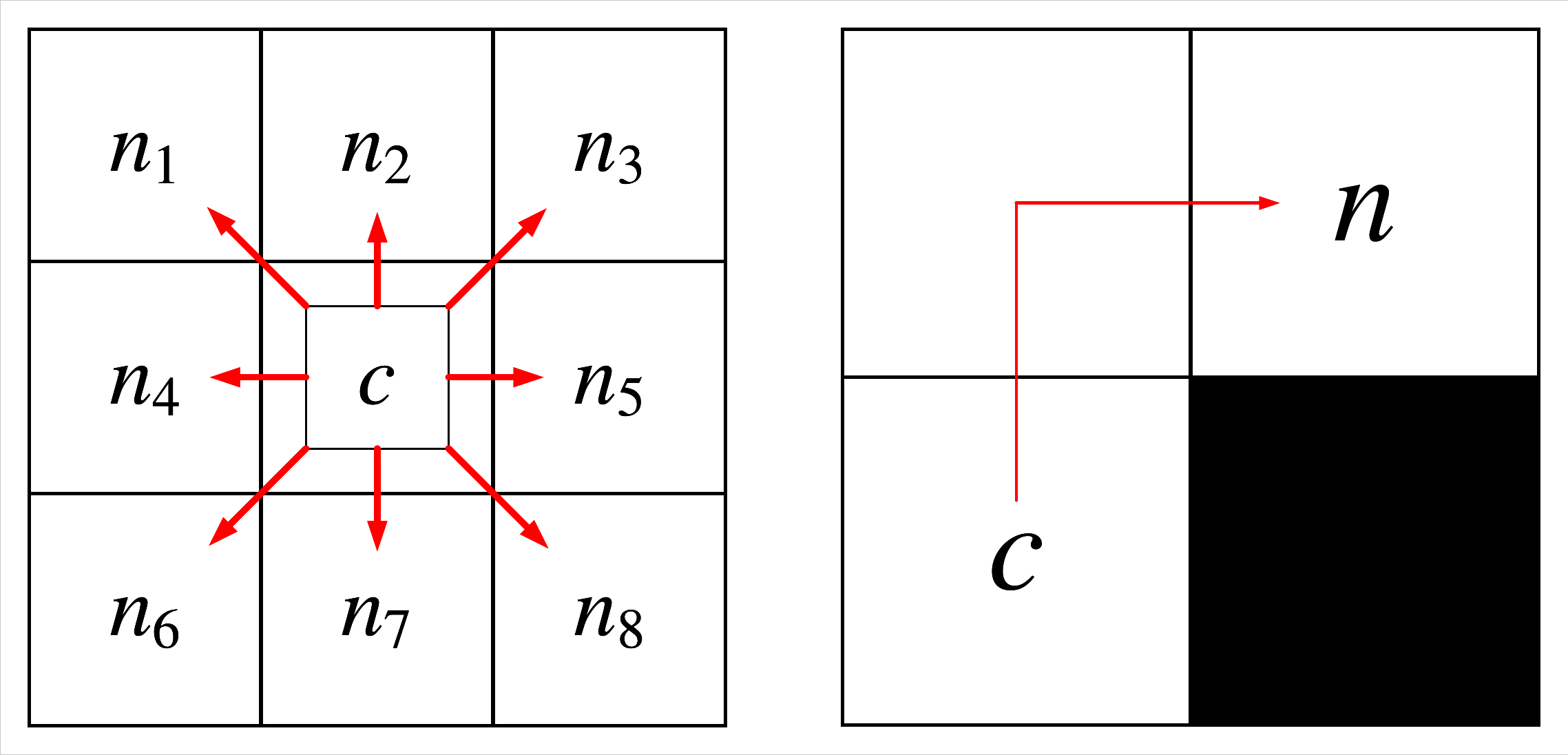}
\caption{The selected path and the correct detour path}
\label{fig:FIG2}
\end{figure}

Path planning algorithm, which adopts an obstacle avoidance strategy, ensures that the planned path neither passes through obstacles nor collides with them. \hyperref[fig:FIG2]{Fig. 4}, which illustrates the agent's movement from grid \textit{c} to grid \textit{n}, shows the path options in 8 directions and the correct detour path.

\begin{figure}[htbp]
\centering
\includegraphics[width=0.36\textwidth]{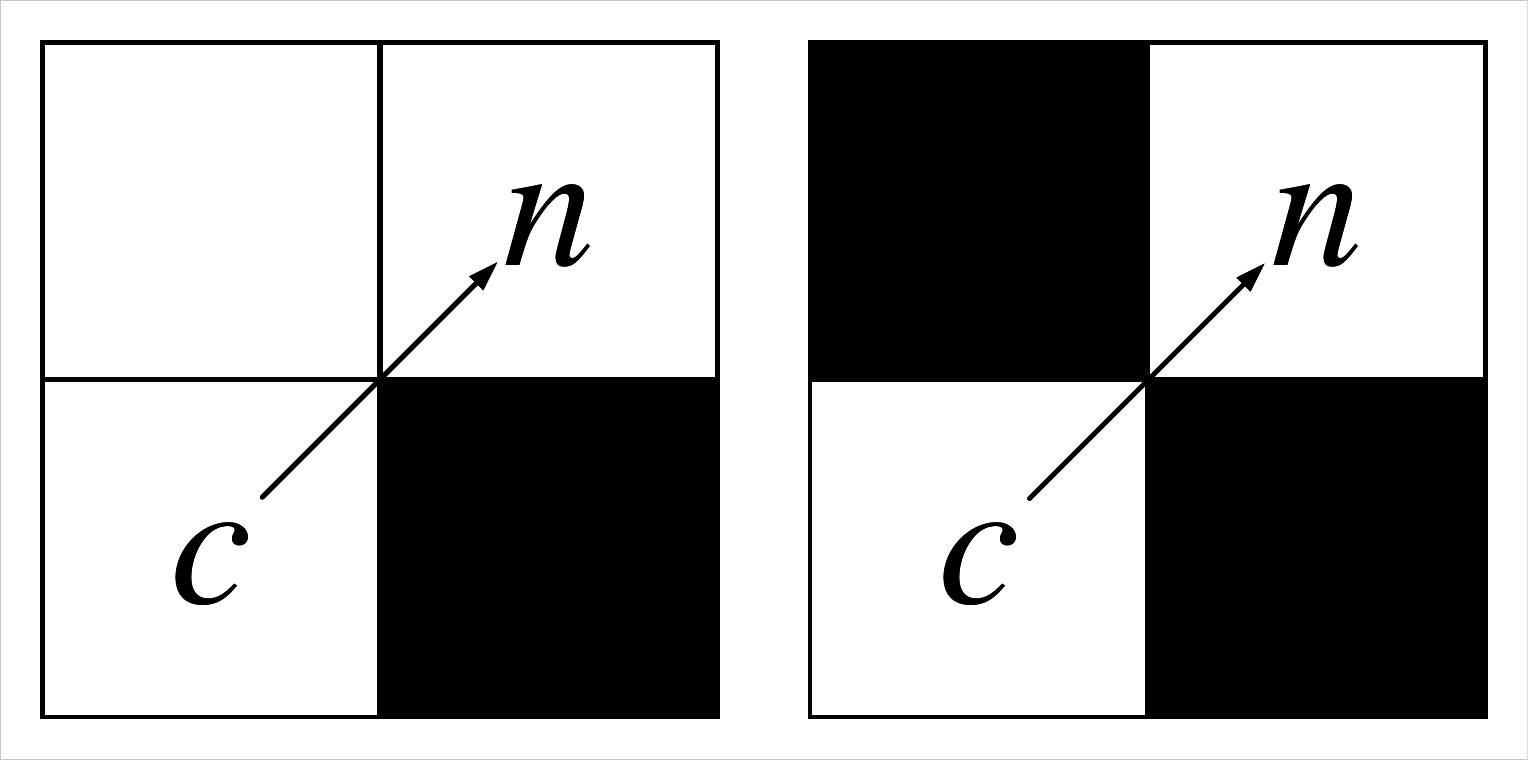}
\caption{Two error paths}
\label{fig:FIG3}
\end{figure}

\hyperref[fig:FIG3]{Fig. 5} illustrates two common error paths, which collide with obstacles and do not meet basic requirements of conditional path constraints.

\section{Methods}\indent

The IQL algorithm introduced in this paper enhances Q-learning for path planning by addressing the inefficiencies of traditional Q-tables. Leveraging the Path Adaptive Collaborative Optimization (PACO) algorithm for Q-table initialization, IQL enables quicker adaptation and more efficient path selection, improving both solution efficiency and accuracy. ACO was chosen over alternatives like Genetic Algorithms and Particle Swarm Optimization for its superior global optimization, adaptability to dynamic environments, and balanced exploration-exploitation. Additionally, ACO’s effectiveness in combinatorial optimization and scalability make it ideal for complex path-planning.

Furthermore, the UCH mechanism is implemented to handle sparse reward functions, reducing exploration risks and enhancing the agent's ability to identify reward criteria. These combined optimizations significantly enhance the IQL algorithm's performance, leading to faster convergence, higher path quality, and improved obstacle avoidance, making it the preferred choice for efficient and adaptable path planning.

Overall, the IQL algorithm, which effectively adapts to the comprehensive needs of path planning efficiency, accuracy, and adaptability, accelerates convergence speed, improves path quality, and strengthens obstacle avoidance.

\subsection{Q-table initialization optimization strategy}\label{subsec2}

\subsubsection{\textbf{PACO algorithm}}\indent

Ant colony optimization (ACO) \cite{ref45} simulates the foraging behavior of real ants. During foraging, ants discharge a volatile substance called pheromones along their path, using its presence and quantity to guide their movement direction. Generally, ants tend to choose paths with more pheromones, forming a positive feedback mechanism: the pheromones on the optimal path increase, while those on other paths gradually decay over time. The ACO algorithm, which simulates this foraging action \cite{ref46}, is a heuristic optimization algorithm with certain advantages in solving combinatorial optimization problems. However, it also has disadvantages such as slow convergence speed, sensitive parameter settings, and a tendency to fall into local optima, which can affect the algorithm's performance and applicability. 

To address these issues, this paper introduces the Path Adaptive Collaborative Optimization (PACO) algorithm to improve ACO algorithm performance. The PACO algorithm is used as the optimization strategy for Q-table initialization, where the best solution obtained by the ACO algorithm serves as the Q-table initialization strategy.

PACO algorithm model is briefly described with the help of \textit{n} random regions. Let \textit{m} ants be placed on \textit{n} random regions, where \textit{n} is the size of the aggregation point; \textit{m} denotes ants’ number; \textit{c} is a set of random areas of this problem; \( \tau _ { i j } ( t )\) is the pheromone on the path region of the random region \textit{i} and \textit{j} at time \textit{t}.

\textbf{Step1: State Transition Guidelines.}

Each ant independently selects the next random region to transfer to based on the pheromones in each path and records the random region that ant \( k \) has traveled in the \( t a b u _k \) table. At time \( t \), the likelihood of ant \( k \) moving from random region \( i \) to random region \( j \), denoted as \( p _ { i j } ^ { k } ( t ) \), is as (4).

\begin{equation}
p_{ij}^k(t) = \left\{ \begin{array}{l}
allowe{d_k} = \left\{ {C - tab{u_k}} \right\}\\
\frac{{{{\left[ {{\tau _{ij}}(t)} \right]}^\alpha }g{{\left[ {{\eta _{ik}}(t)} \right]}^\beta }}}{{\sum\limits_{s \in allowe{d_k}} {{{\left[ {{\tau _{is}}(t)} \right]}^\alpha }g{{\left[ {{\eta _{is}}(t)} \right]}^\beta }} }}\;j \in allowe{d_k}\\
0\quad \quad \quad \quad \quad \quad \quad \quad \quad \quad j \notin allowe{d_k}
\end{array} \right.
\end{equation}
Among them, \( \alpha \) is the information heuristic, reflecting the influence of pheromones on the path region of the ant, and \( \beta \) is the expectation heuristic, indicating the impact of the path area on the ant. \( a l l o w e d _k \) is the random region set that can be selected when the ant \( k \) moves next.

\textbf{Step2: Pheromone updates.}

In the PACO algorithm, the probability of choosing a path is proportional to the pheromone concentration on that path. Over time, pheromone levels decrease according to a defined volatility coefficient. Traditional ant colony algorithms use a fixed value for this coefficient. If set incorrectly, the colony may prematurely converge on certain paths, resulting in local optima, thereby hindering the algorithm's ability to find the global best solution and slowing down convergence.

To address this limitation, the PACO algorithm incorporates a mechanism for dynamically adjusting the pheromone volatility factor. Instead of being a static constant, the pheromone volatility factor \( \rho \) decreases gradually with each iteration. This adjustment maintains path diversity and enhances convergence speed, as demonstrated in (5).

\begin{equation}
\rho ( t ) = \frac { \lambda _ { 1 } } { ( 1 + e ^ { \frac { \lambda t } { 3 m } } ) }
\end{equation}
where \( \lambda _ { 1 }\) represents the adjustment coefficient of the pheromone volatilization factor, \textit{Nc} represents the number of the current iteration, and \textit{m} is the ant’s number.

The pheromone update formula in the PACO algorithm is shown in (6).

\begin{equation}
\left\{ \begin{array}{l}
{\tau _{ij}}(t + 1) = (1 - \rho (t)) \bullet {\tau _{ij}}(t) + \Delta {\tau _{ij}}(t)\\
\Delta {\tau _{ij}}(t) = \sum\limits_{k = 1}^m {\Delta \tau _{ij}^k(t),\Delta \tau _{ij}^k(t) = \left\{ \begin{array}{l}
\frac{Q}{{{L_k}}},(i,j) \in {L_k}\\
0\;\;\;,others
\end{array} \right.} 
\end{array} \right.
\end{equation}

Table 2 shows the symbols and their meanings in (6).

\begin{table}[htbp]
  \centering
  \caption{Symbols of pheromone update formulas in the PACO algorithm}
   \centering
  \begin{tabular}{c c}
    \hline
    Symbols & Symbolic meanings \\
    \hline
    \(\Delta {\tau _{ij}}(t)\) & The pheromone increment on path (\textit{i},\textit{j}) at time \textit{t} \\
    \textit{k} & The \textit{k}-th ant \\
    \textit{m} & Total ants number \\
    \textit{Q} & Initial intensity of pheromone increment (fixed constant) \\
    \({L_k}\) & The \textit{k}-th ant searches for the total length of the path at time \textit{t} \\
    \hline
  \end{tabular}
\end{table}

\begin{figure}[htbp]
\centering
\includegraphics[width=0.8\textwidth]{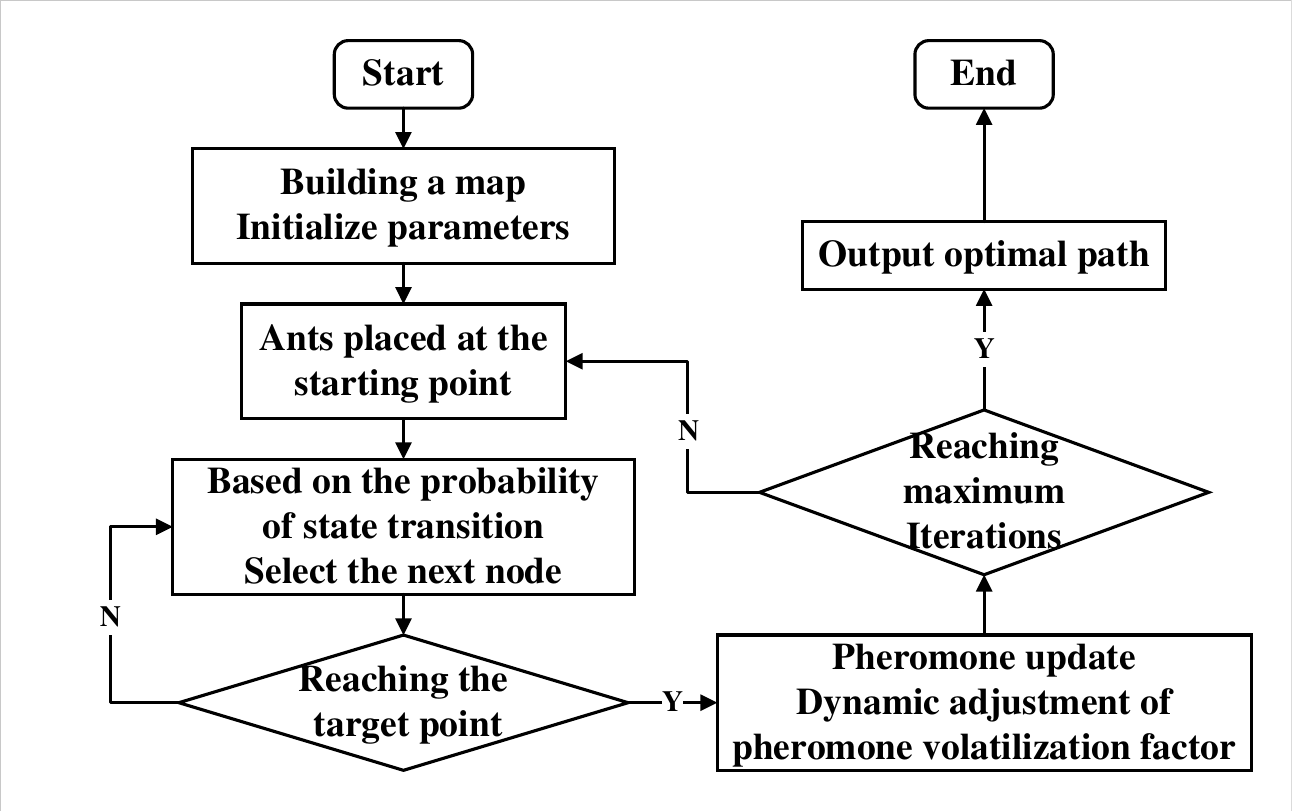}
\caption{Flow chart of the PACO algorithm}
\label{fig:FIG6}
\end{figure}

The PACO algorithm draws on the feeding behavior characteristics of ants in nature to form a distributed intelligence system, and the pheromone update mechanism is improved by introducing the volatilization coefficient of the pheromone to avoid the problem of ants falling into a local optimum, as shown in \hyperref[fig:FIG6]{Fig. 6}.

\subsubsection{\textbf{Q-table initialization operation}}\indent

\begin{figure}[htbp]
\centering
\includegraphics[width=0.8\textwidth]{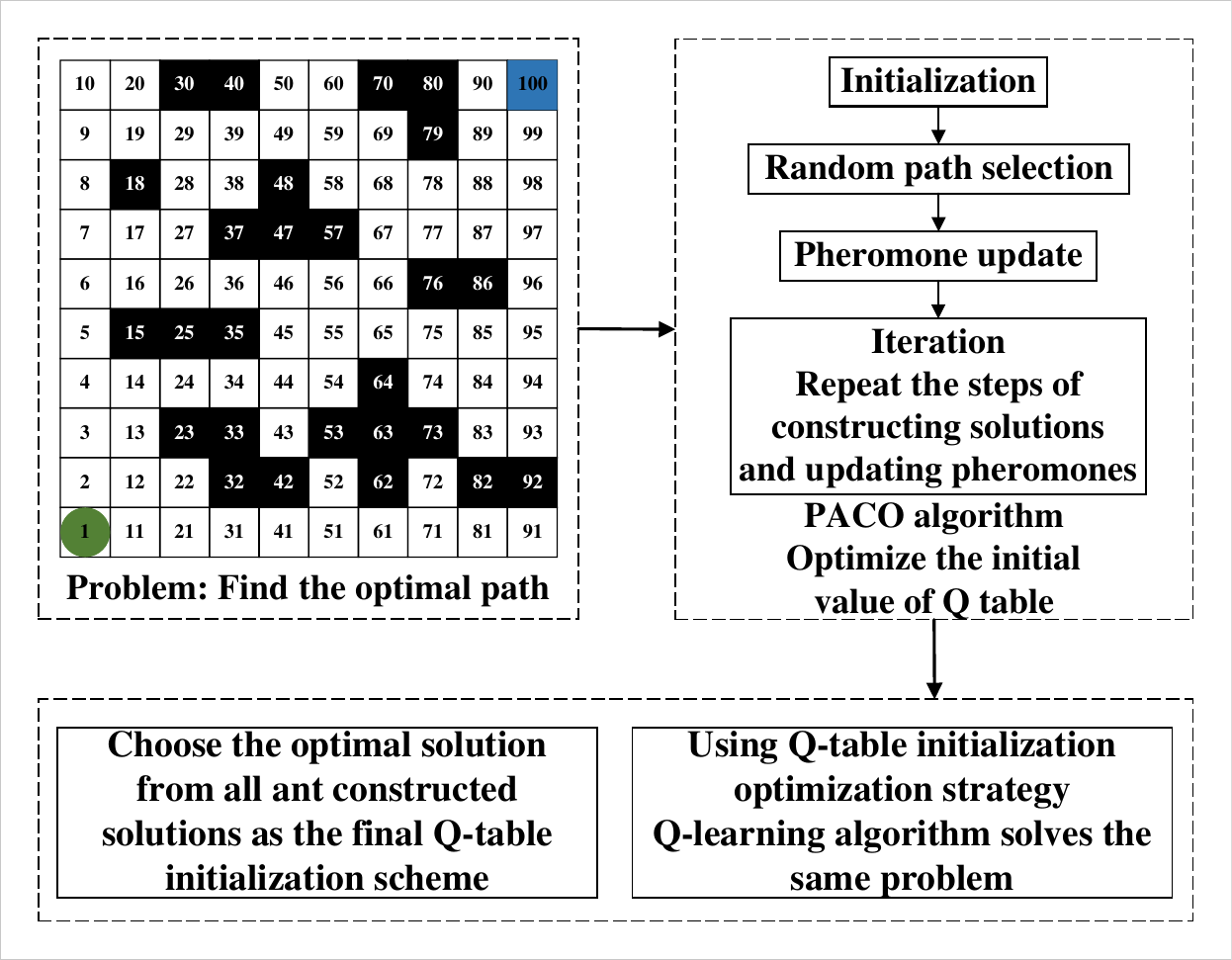}
\caption{Initialization optimization policy process of the Q-table}
\label{fig:FIG7}
\end{figure}

Q-table initialization optimization strategy basis of the PACO algorithm involves these steps, which could be summarized to describe the process. \hyperref[fig:FIG7]{Fig. 7} shows the Q-table initialization optimization strategy process based on the PACO algorithm.

\textbf{Step1: Problem definition.} Identify the problem of finding the optimal path that needs to be solved.

\textbf{Step2: PACO algorithm parameter settings.} Set the parameters of the PACO algorithm, such as the total number of ants, the initial intensity of pheromone increment, the expected heuristic, and the pheromone volatilization rate.

\textbf{Step3: Colony initialization.} Randomly place ants at the start of the map, with each ant representing a potential solution.

\textbf{Step4: Iterative process.} Select the next node of the path based on the state transition probability and determine whether the target point is reached.

\textbf{Step5: Pheromone updates.} The path diversity is ensured by introducing the mechanism of dynamic regulation of pheromone volatilization factors, and the convergence rate is improved.

\textbf{Step6: Output the result.} After the algorithm is terminated, the best solution found is output as the initial value of the Q table.

\textbf{Step7: Q-table initialization.} The initial utility value of the ant taking different actions in each state is the IQL algorithm's initial value. This step is a vital part of the optimization strategy.

\textbf{Step8: Algorithm performance evaluation.} The results are evaluated to evidence the effectiveness of the optimization strategy.

\subsection{Reward function optimization}\indent

The reward function, a crucial component of Q-learning algorithms, determines the reward an agent receives from interacting with the environment. By optimizing the reward function, the agent can learn an effective strategy more quickly. In this section, we optimize the reward function by introducing a Utility-Controlled Heuristic (UCH) mechanism, which significantly enhances the algorithm's performance.

The UCH mechanism dynamically adjusts the reward parameter and changes the distance calculation method. This mechanism helps the reward function develop an effective strategy faster and in real time. Table 3 uses Euclidean distance as an example to compare the optimized and original reward functions.

\begin{table}[htbp]
  \centering
  \caption{Comparison between the original reward function and the optimized reward function (distance calculation method: Euclidean distance)}
   \centering
  \begin{tabular}{c c}
    \hline
    Meaning & Formula \\
    \hline
    Raw reward function & \(r(s,a) =  - \sqrt {{{({x_s} - {x_s}^{\prime \prime })}^2} + {{({y_s} - {y_s}^{\prime \prime })}^2}} \) \\
    Parameter formulas & \( \mu ( t ) = \mu _ { 0 } \cdot \frac { 1 } { \pi + \pi \cdot e ^ { - t } }\) \\
    Optimized reward function & \({r^O}(s,a) =  - \mu (t){\kern 1pt} {\kern 1pt} r(s,a)\) \\
    \hline
  \end{tabular}
\end{table}

In this paper, we compare two distance calculation methods with the Euclidean distance method used in the original algorithm.

\textbf{(1) Chebyshev distance}

In mathematics, Chebyshev distance \cite{ref47}, also known as the \( L \infty\) norm, is regarded as a measure within a vector space. The distance is determined by calculating the absolute difference between two points in each coordinate dimension and taking the maximum value from these differences.

Chebyshev distance is based on the concept of a consistent norm (or supremum norm) and is classified as an injective metric space.

This paper only discusses the Manhattan distance in a two-dimensional plane. Let the Chebyshev distance between the two points on the plane be \( A ( x _ { s } , y _ { s } )\) and \( B ( x _ { s } ^ { \prime \prime } , y _ { s } ^ { \prime \prime } )\) , and the Chebyshev distance of the two points \textit{AB} as (7).

\begin{equation}
d _ { A B } ^ { Q } = \max ( | x _ { s } - x _ { s } ^ { \prime \prime } | , | y _ { s } - y _ { s } ^ { \prime \prime } | )
\end{equation}

Table 4 compares the optimized reward function with the original reward function using the Chebyshev distance as an example.

\begin{table}[htbp]
  \centering
  \caption{Comparison between the original reward function and the optimized reward function (distance calculation method: Chebyshev distance)}
   \centering
  \begin{tabular}{c c}
    \hline
    Meaning & Formula \\
    \hline
    Raw reward function & \( r _ { q } ( s , a ) = - \max ( | x _ { s } - x _ { s } ^ { \prime \prime } | , | y _ { s } - y _ { s } ^ { \prime \prime } | )\) \\
    Parameter formulas & \( \mu ( t ) = \mu _ { 0 } \cdot \frac { 1 } { \pi + \pi \cdot e ^ { - t } }\) \\
    Optimized reward function & \( r ^ { Q } ( s , a ) =  \rho ( t ) \cdot \ r _ { q } ( s , a )\) \\
    \hline
  \end{tabular}
\end{table}

\textbf{(2) Manhattan distance}

In Manhattan neighborhoods, the driving distance from one intersection to another is not a straight-line distance between two points, and this actual distance is the "Manhattan distance \cite{ref48}". For this reason, Manhattan distance is also known as "taxi distance" or "city block distance".

This paper only discusses the Manhattan distance in a two-dimensional plane. Let the two points on the plane be \( A ( x _ { s } , y _ { s } )\) and \( B ( x _ { s } ^ { \prime \prime } , y _ { s } ^ { \prime \prime } )\) , and the Manhattan distance between the two points \textit{AB} as (8).

\begin{equation}
d _ { A B } ^ { M } = | x _ { z } - x _ { z } ^ { \prime \prime } | + | y _ { z } - y _ { z } ^ { \prime \prime } |
\end{equation}

Table 5 below takes the Manhattan distance as an example to show the comparison between the optimized reward function and the original reward function.

\begin{table}[htbp]
  \centering
  \caption{Comparison between the original reward function and the optimized reward function (distance calculation method: Manhattan distance)}
   \centering
  \begin{tabular}{c c}
    \hline
    Meaning & Formula \\
    \hline
    Raw reward function & \( r _ { m } ( s , a ) = - \max ( | x _ { s } - x _ { s } ^ { \prime \prime } | , | y _ { s } - y _ { s } ^ { \prime \prime } | )\) \\
    Parameter formulas & \( \mu ( t ) = \mu _ { 0 } \cdot \frac { 1 } { \pi + \pi \cdot e ^ { - t } }\) \\
    Optimized reward function & \( r ^ { M } ( s , a ) =  \rho ( t ) \cdot \ r _ { m } ( s , a )\) \\
    \hline
  \end{tabular}
\end{table}

\hyperref[fig:FIG8]{Fig. 8} below illustrates the difference and connection between the Euclidean distance, the Manhattan distance, and the equivalent Manhattan distance.

\begin{figure}[htbp]
\centering
\includegraphics[width=0.36\textwidth]{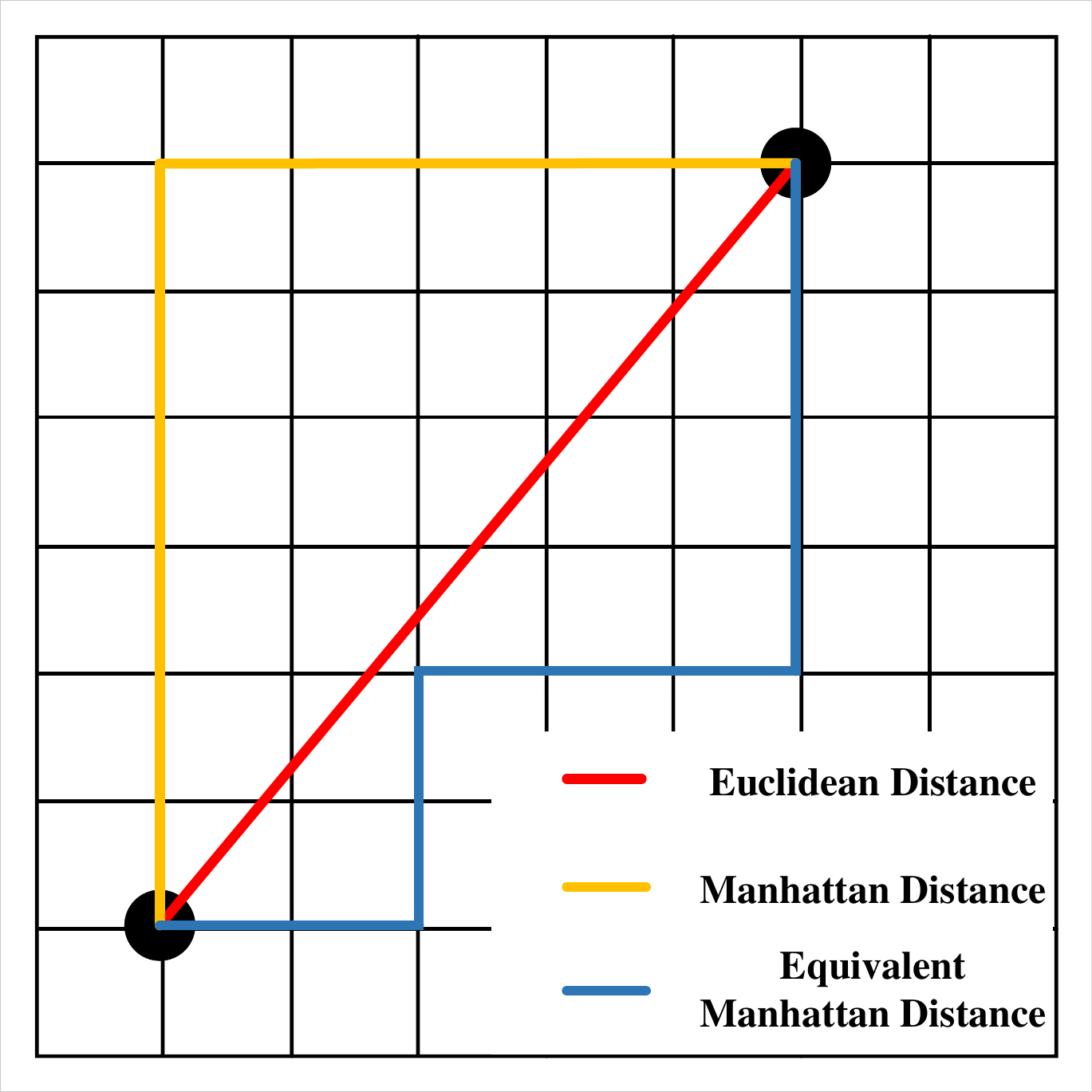}
\caption{Euclidean distance, Manhattan distance, and equivalent Manhattan distance}
\label{fig:FIG8}
\end{figure}

\subsection{Evaluation indicators}\indent

Algorithm evaluation metrics quantitatively measure performance, including learning speed, stability, and final outcomes. These metrics provide a comprehensive assessment of the algorithm's practical effectiveness. Therefore, three evaluation indicators are used to assess the IQL algorithms.

\textbf{(1) Trials number with the same number of iterations X. (\( \eta ( t )\))} 

With the same number of iterations, the algorithm requires fewer experiments to reach a steady state, indicating improved performance.

\textbf{(2) The number of trials with a standard deviation of 0. (\( d(t) \))} 

\( d(t) \) is a statistic that measures the degree of dispersion in a set of numerical distributions. If the optimized algorithm achieves a standard deviation of 0 with fewer tests, it indicates improved measurement stability and reduced systematic error, and enhanced data reliability.

\textbf{(3) The expected return after the algorithm's convergence. (\( e(t) \))} 

\( e(t) \) refers to the average cumulative return an agent receives from the initial state when following a strategy. The agent can better utilize the learned strategy to maximize cumulative rewards if the algorithm provides a higher \( e(t) \) in the initial state.

Therefore, the formula \( J ( t )\) that defines the performance evaluation index of the experiment in this paper is as (9).

\begin{equation}
J(t) = \frac{{\frac{{\eta (1) - \eta (2)}}{{\eta (1)}} + \frac{{d(1) - d(2)}}{{d(1)}} + \frac{{e(1) - e(2)}}{{e(2)}}}}{3} \times 100\% 
\end{equation}

Table 6 illustrates the meaning of the symbols in (9). The \textit{i}'s in Table 6 are 1 or 2.

\begin{table}[htbp]
  \centering
  \caption{Description of the symbol of \( J ( t )\)}
   \centering
  \begin{tabular}{c c p{0.5\textwidth}}
    \hline
    Symbols & Symbolic meaning \\
    \hline
    1 & The original algorithm \\
    2 & The improved algorithm \\
    \( \eta ( i )\) & Trials number for the algorithm to reach steady state \\
    \( d ( i )\) & The algorithm's trials number with standard deviation to 0 \\
    \( e ( i )\) & The algorithm accumulates expected returns \\
    \hline
  \end{tabular}
\end{table}

\section{Experiments}
\subsection{Simulation environment}\indent

Below are three types of grid environments used in this article.

\begin{figure}[htbp]
\centering
\includegraphics[width=0.42\textwidth]{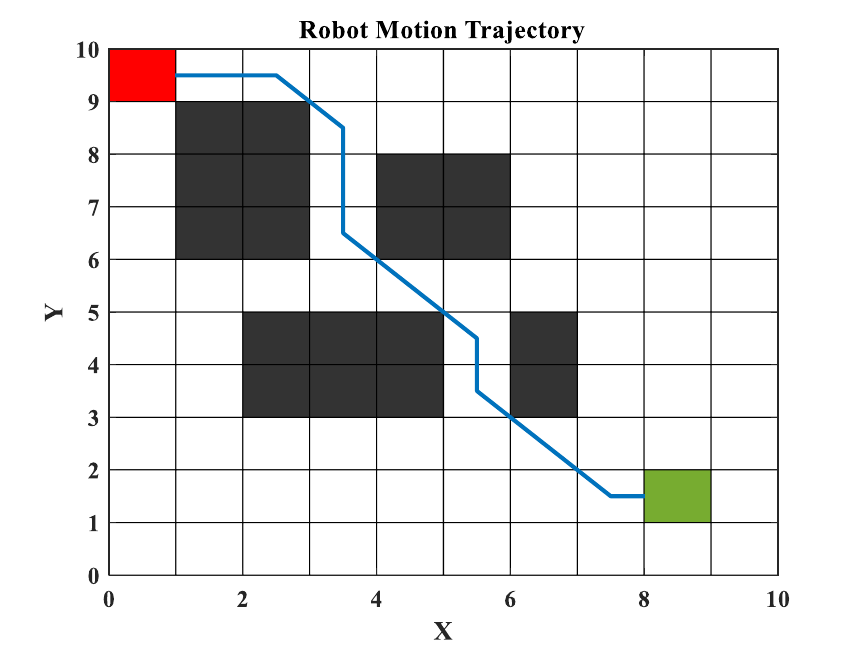}
\caption{10*10 raster map environment (S10)}
\label{fig:FIG9}
\end{figure}

The first raster map, shown in \hyperref[fig:FIG9]{Fig. 9} (S10), is a low-dimensional, simple environment with a few obstacles, where the robot can quickly find the optimal route between the start and endpoints. This setup is used to test the basic functionality and learning efficiency of the IQL algorithm.

\begin{figure}[htbp]
\centering
\includegraphics[width=0.42\textwidth]{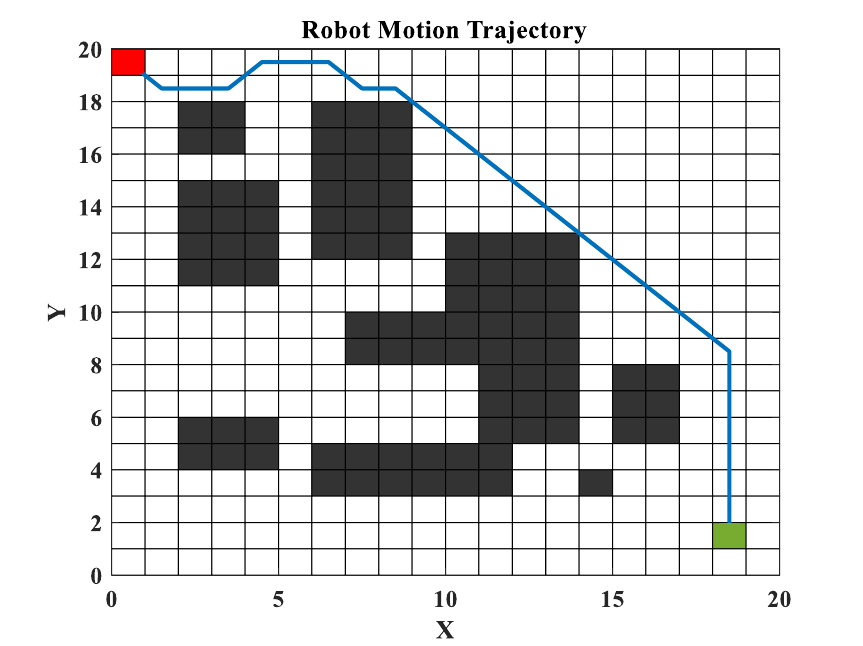}
\caption{20*20 raster map environment (S20)}
\label{fig:FIG10}
\end{figure}

The second environment, i.e., the map of \hyperref[fig:FIG10]{Fig. 10} (S20), is of medium dimensionality and has a more complex structure, including more black grids. This environment allows for the testing of the route planning ability and computational efficiency of the IQL algorithms in the face of more complex situations.

\begin{figure}[htbp]
\centering
\includegraphics[width=0.42\textwidth]{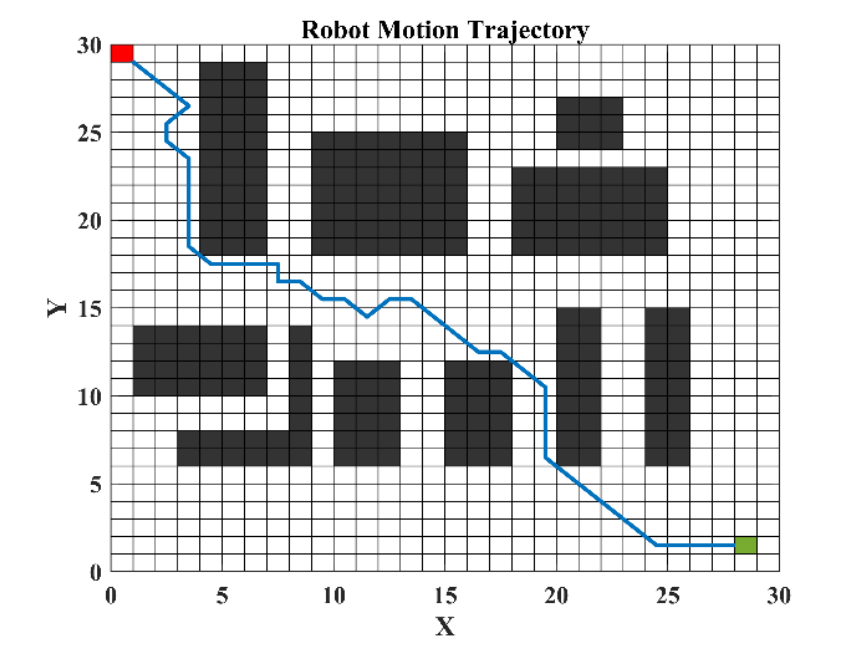}
\caption{30*30 raster map environment (S30)}
\label{fig:FIG11}
\end{figure}

The third environment, the map in \hyperref[fig:FIG11]{Fig. 11} (S30), is a high-dimensional complex map designed with a large number of obstacles. This environment simulates a challenging navigation task in the real world. It is used to estimate the performance of the IQL algorithm in dealing with elaborate environments, especially in planning speed and obstacle avoidance.

Three MATLAB-designed raster map environments of varying complexity were implemented to evaluate the IQL algorithm's performance in route planning. These environments simulate real-world robot path planning, with each cell representing a potential robot position. Black grids denote obstacles, while white grids represent navigable paths, providing a clear framework for algorithm development and testing.

A series of simulation experiments in three different dimensional raster map environments were conducted to compare the performance of the IQL algorithm with traditional Q-learning and other algorithms. This comprehensive evaluation highlights the practical effects and potential application value of the proposed optimization methodology for path planning problems. Basic parameter settings chosen for this experiment are presented in Table 7.

\begin{table}[htbp]
  \centering
  \caption{IQL algorithms parameter settings}
   \centering
  \begin{tabular}{c c p{0.3\textwidth}}
    \hline
    Name of the parameters involved & Value or range \\
    \hline
    Learning rate (\( \alpha\)) & 0.3 \\
    Discount factor (\( \gamma\)) & 0.95 \\
    Q-Table Initializations & 4.5598/11.5598/9.3397/7.5598 \\
    \( \rho _ { 0 }\)in the optimized reward function & 0.156/0.0156/0.016/0.01056 \\
    Convergence target & 0.25 \\
    Convergence Iterations Average & 10 \\
    \hline
  \end{tabular}
\end{table}

The algorithm runtime environment and computer configuration in this document are shown in Table 8.

\begin{table}
\caption{The algorithm runtime environment and computer configuration\label{tab:table}}
\centering
\begin{tabular}{c c}
\hline
Parameter & Configuration\\
\hline
Device name & LAPTOP-A3S8EAVD\\
Processor & Intel(R) Core(TM) i5-8265U CPU @ 1.60GHz\\
With RAM & 8.00 GB (7.85 GB available)\\
System type & 64-bit operating system, x64-based processor\\
Simulation software & MATLAB R2022a\\
\hline
\end{tabular}
\end{table}

\subsection{Algorithm validation}\indent

In this section, the following four algorithms are tested in each of the above three environments:

\textit{a}).  Q-learning algorithm.

\textit{b}).  A Q-learning algorithm for optimizing the initial values of Q-tables.

\textit{c}).  Q-learning algorithm with improved reward function.

\textit{d}).  IQL algorithm.

After preliminary experiments and parameter tuning, the IQL algorithm achieved optimal learning and path planning with a Q-table initialized at 9.3397, \( \rho _ { 0 }\) set to 0.016 in the optimized reward function, and Chebyshev distance for computation. These parameters were used in all subsequent experiments. The following tables (Table 9, Table 10 and Table 11) present the results for a Q-table initialized at 9.3397, \( \rho _ { 0 }\) set to 0.016, comparing Chebyshev and Manhattan distance calculations.

\begin{table}[htbp]
  \centering
  \caption{Experimental results of four algorithms in a 10*10 raster map environment}
   \centering
  \begin{tabular}{c c c c c p{0.5\textwidth}}
    \hline
    \centering
    Algorithm name & Distance calculation &  \( \eta ( t )\) & \( d ( t )\) & \( e ( t )\) \\
    \hline
    \textit{a}) & Euclidean distance & 157 & 147 & 2543.26 \\
    \textit{b}) & Euclidean distance & 143 & 133 & 2680.80 \\
    \textit{c}) & Chebyshev distance & 137 & 127 & 2674.77 \\
    \textit{c}) & Manhattan distance & 145 & 135 & 2666.29 \\
    \textit{d}) & Chebyshev distance & 133 & 123 & 2726.95 \\
    \textit{d}) & Manhattan distance & 142 & 132 & 2686.52 \\
    \hline
  \end{tabular}
\end{table}

\begin{table}[htbp]
  \centering
  \caption{Experimental results of four algorithms in a 20*20 raster map environment}
   \centering
  \begin{tabular}{c c c c c p{0.5\textwidth}}
    \hline
    \centering
    Algorithm name & Distance calculation &  \( \eta ( t )\) & \( d ( t )\) & \( e ( t )\) \\
    \hline
    \textit{a}) & Euclidean distance & 486 & 476 & 8794.61 \\
    \textit{b}) & Euclidean distance & 430 & 420 & 9210.92 \\
    \textit{c}) & Chebyshev distance & 426 & 416 & 9156.23 \\
    \textit{c}) & Manhattan distance & 432 & 422 & 9150.13 \\
    \textit{d}) & Chebyshev distance & 376 & 366 & 9594.86 \\
    \textit{d}) & Manhattan distance & 408 & 398 & 9395.92 \\
    \hline
  \end{tabular}
\end{table}

\begin{table}[htbp]
  \centering
  \caption{Experimental results of four algorithms in a 30*30 raster map environment}
   \centering
  \begin{tabular}{c c c c c p{0.5\textwidth}}
    \hline
    \centering
    Algorithm name & Distance calculation &  \( \eta ( t )\) & \( d ( t )\) & \( e ( t )\) \\
    \hline
    \textit{a}) & Euclidean distance & 779 & 769 & 14096.71 \\
    \textit{b}) & Euclidean distance & 732 & 722 & 14526.66 \\
    \textit{c}) & Chebyshev distance & 725 & 715 & 14712.74 \\
    \textit{c}) & Manhattan distance & 729 & 719 & 14549.21 \\
    \textit{d}) & Chebyshev distance & 707 & 697 & 14849.47 \\
    \textit{d}) & Manhattan distance & 713 & 703 & 14743.75 \\
    \hline
  \end{tabular}
\end{table}

Overall, the IQL algorithm performance is significantly improved for simple scenes in low dimensions and complex maps in higher dimensions.

In the 10*10 and 20*20 raster maps, the IQL algorithm using Chebyshev distance significantly reduces the trials needed for convergence—by 15.29\% and 22.63\%, respectively—while also increasing the expected return by 7.23\% and 9.11\%. The algorithm performs even better in the 20*20 environment, demonstrating its effectiveness in handling larger, more complex problems. In the 30*30 map, although improvements are smaller, the algorithm still reduces trials by 9.24\% and increases expected returns by 5.34\%. Table 12 shows the optimized algorithm's performance compared to the original algorithm.

\begin{table}[htbp]
  \centering
  \caption{Performance comparison between the optimized algorithm and the original algorithm}
   \centering
  \begin{tabular}{c c c c c c}
    \hline
    \centering
    Algorithm & Distance & Evaluation & 10*10 & 20*20 & 30*30 \\
    \hline
    \textit{d}) & Chebyshev & \( \eta ( t )\) & 15.29\% & 22.63\% & 9.24\% \\
    \textit{d}) & Chebyshev & \( d ( t )\) & 16.33\% & 23.11\% & 9.36\% \\
    \textit{d}) & Chebyshev & \( e ( t )\) & 7.23\% & 9.11\% & 5.34\% \\
    \textit{d}) & Manhattan & \( \eta ( t )\) & 9.55\% & 16.05\% & 8.47\% \\
    \textit{d}) & Manhattan & \( d ( t )\) & 10.21\% & 16.39\% & 8.58\% \\
    \textit{d}) & Manhattan & \( e ( t )\) & 5.63\% & 6.84\% & 4.59\% \\
    \hline
  \end{tabular}
\end{table}

Chebyshev distance is chosen over Manhattan distance for its more accurate modeling of movement costs in grid-based environments, where diagonal movements are equally important as horizontal or vertical ones. This choice enhances the efficiency and accuracy of the IQL algorithm across various map dimensions, with observed performance improvements further validating its effectiveness.

\begin{figure}[htbp]
\centering
\includegraphics[width=0.6\textwidth]{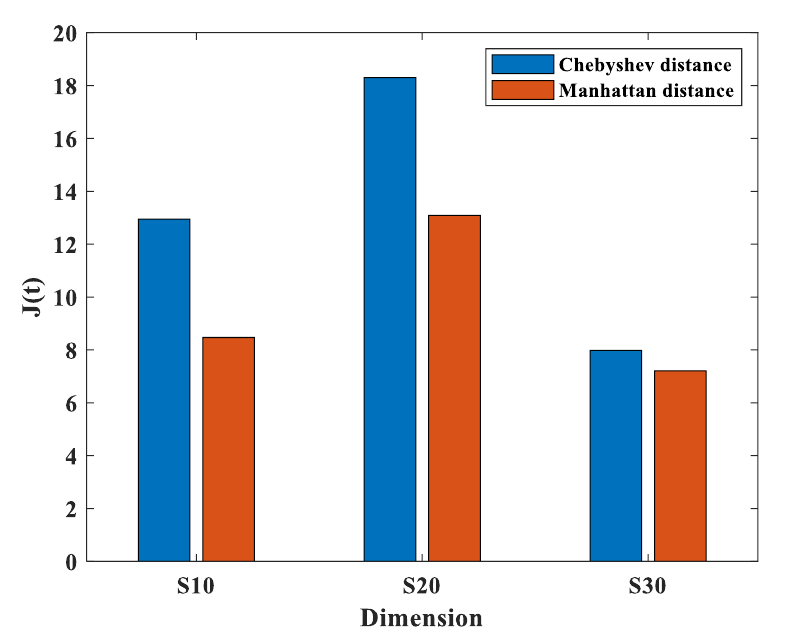}
\caption{Comparison of Algorithm Performance Using Different Distance Calculation Methods}
\label{fig:FIG12}
\end{figure}

\subsection{Algorithm comparison}\indent

This section highlights the advantages of the IQL algorithm by comparing it with the FIQL, PP-Q-Learning-based CPP (PP-QL-based CPP), DFQL, and QMABC algorithms. The comparison follows these steps.

\textbf{Step1: Clarify the purpose of the comparison.}

This program aims to compare the performance of different algorithms in 10*10 and 20*20 raster environments. By examining these scenarios, we can understand the variation in efficiency and effectiveness of three algorithms across different raster map sizes.

\textbf{Step2: Selection of benchmarking algorithms.}

This section compares the IQL algorithm with the FIQL, PP-QL-based CPP, DFQL, and QMABC algorithms. These comparison algorithms are all variants of Q-learning that use different approaches to enhance learning speed and performance. This comparison demonstrates the superiority of the IQL algorithm.

\textbf{Step3: Algorithm performance evaluation metrics.}

Evaluation metrics are crucial for measuring algorithm performance. We will use three metrics from Section Methods: the number of trials X under consistent iterations (\( \eta ( t )\)), the number of trials where the standard deviation reaches 0 (\( d ( t )\)), and the expected return after convergence (\( e ( t )\)). All algorithms must be run under the same conditions and tested on the same dataset. Additionally, the performance of each metric should be recorded and statistically analyzed to identify the best-performing algorithm for the given scenario.

\textbf{Step4: Comparative analysis of algorithms.}

Table 13 and Table 14 compared the experimental results of the Q-learning algorithm with the IQL algorithm, the FIQL algorithm, the PP-QL-based CPP algorithm, the DFQL algorithm, and the QMABC algorithm.

\begin{table}[htbp]
  \centering
  \caption{The experimental results of the three algorithms are shown and compared (10*10)}
   \centering
  \begin{tabular}{c c c c c p{0.5\textwidth}}
    \hline
    \centering
    Algorithm name & \( \eta ( t )\) & \( d ( t )\) & \( e ( t )\) \\
    \hline
    \textit{a}). Q-learning algorithm & 165 & 155 & 2543.26 \\
    \textit{d}). IQL algorithm & 133 & 123 & 2726.95 \\
    FIQL algorithm & 144 & 134 & 2688.18 \\
    PP-QL-based CPP algorithm & 141 & 131 & 2694.95 \\
    DFQL algorithm & 149 & 139 & 2597.78 \\
    QMABC algorithm & 146 & 136 & 2622.12 \\
    \hline
  \end{tabular}
\end{table}

\begin{table}[htbp]
  \centering
  \caption{The experimental results of the three algorithms are shown and compared (20*20)}
   \centering
  \begin{tabular}{c c c c c p{0.5\textwidth}}
    \hline
    \centering
    Algorithm name & \( \eta ( t )\) & \( d ( t )\) & \( e ( t )\) \\
    \hline
    \textit{a}) Q-learning algorithm & 486 & 476 & 8794.61 \\
    \textit{d}) IQL algorithm & 376 & 366 & 9594.86 \\
    FIQL algorithm & 406 & 396 & 9297.78 \\
    PP-QL-based CPP algorithm & 392 & 382 & 9322.12 \\
    DFQL algorithm & 427 & 417 & 9367.78 \\
    QMABC algorithm & 415 & 405 & 9372.12 \\
    \hline
  \end{tabular}
\end{table}

In the 10*10 and 20*20 raster map environments, the IQL algorithm demonstrated significant performance advantages. For example, in the 10*10 grid, compared to traditional Q-learning, the trials required to reach a steady state with a standard deviation of 0 were reduced by 20.65\%, 13.55\%, 15.48\%, 10.32\%, and 12.26\% for the IQL, FIQL, PP-QL-based CPP, DFQL, and QMABC algorithms, respectively. In the 20*20 grid, the reductions were 20.11\%, 16.81\%, 19.75\%, 12.39\%, and 14.92\%, respectively. Among the five algorithms, IQL exhibited the fastest convergence speed and the highest stability, enabling more effective task learning and execution, and achieving optimal performance.

\begin{figure}[htbp]
\centering
\includegraphics[width=1.0\textwidth]{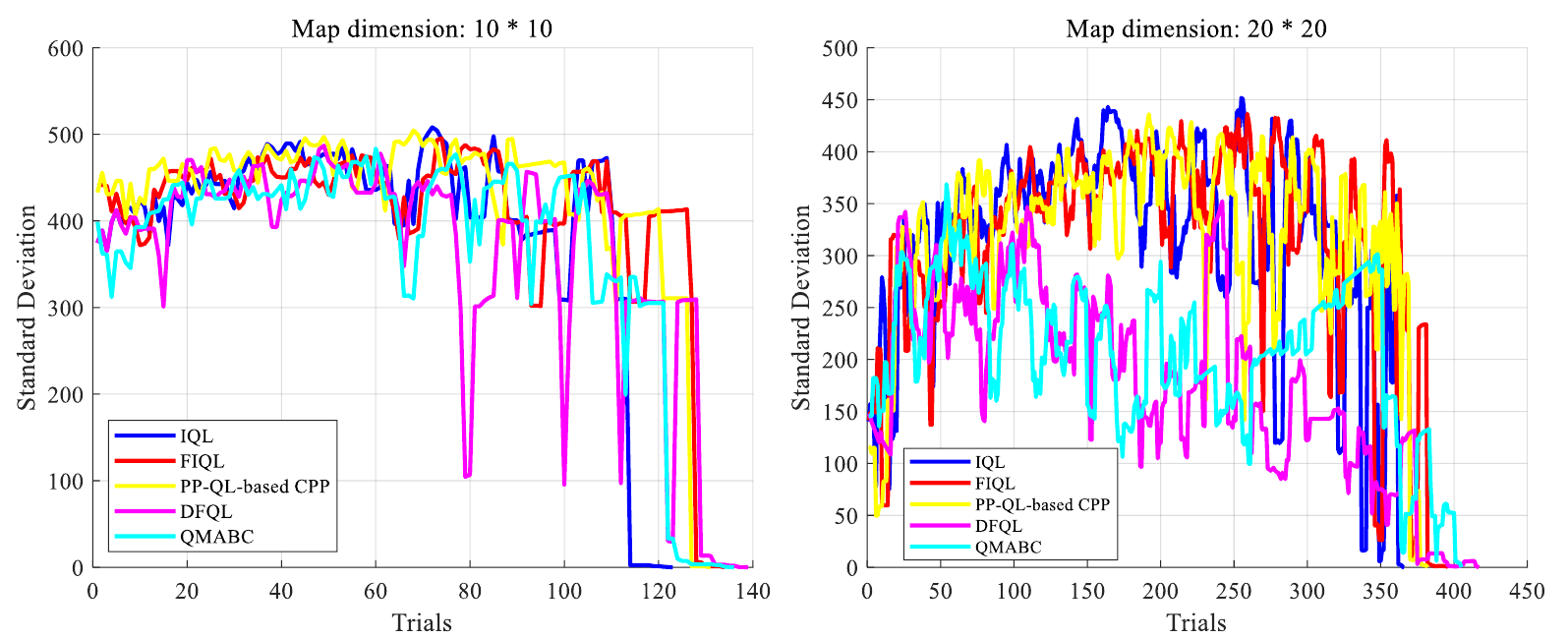}
\caption{Comparison of the number of trials in which the standard deviation of the five different algorithms reaches 0 for two types of raster environments}
\label{fig:FIG12}
\end{figure}

The graphs illustrate the results of experiments in two different raster environments, comparing the FIQL algorithm, the PP-QL-based CPP algorithm, and the improved IQL algorithm. This section focuses on the number of trials where the standard deviation reaches 0, a key stability and convergence speed metric. A smaller standard deviation indicates more stable data; a standard deviation 0 means complete stability. As shown in \hyperref[fig:FIG12]{Fig. 13}, the IQL algorithm converges faster than the other two algorithms in both raster environments. This quick convergence is valuable for practical applications as it reduces the time needed for learning and planning. The IQL algorithm not only converges faster but also achieves better final performance.

The IQL algorithm enhances learning efficiency and task performance within the same environment, boosting overall performance, which is crucial for tackling more complex and larger-scale problems. In summary, its performance in a raster environment demonstrates its effectiveness in improving learning efficiency, stability, and outcomes, offering valuable insights for future algorithm design and optimization in similar settings.

\subsection{Discussion}\indent

Our work presents an optimized algorithm that enhances Q-table initialization and refines reward functions, enabling faster and more efficient strategy development in complex environments by minimizing unnecessary trial and error. This approach shows broad applicability across various raster map environments, consistently improving the IQL algorithm's ability to navigate high-dimensional maps with numerous obstacles. Experimental results confirm the algorithm's stability and reliability, validating its effectiveness in path planning through both theoretical and empirical analysis.

The IQL algorithm shows enhanced performance across all map dimensions. In low-dimensional maps, it efficiently guides the agent to avoid obstacles and find the optimal path, significantly outperforming traditional Q-learning. Even in more complex 30*30 maps, the algorithm reduces trials needed for convergence and increases expected returns, consistently providing better stability, faster convergence, and more effective pathfinding across all tested dimensions.

These findings have significant implications for practical applications, particularly in the operation of autonomous robots or vehicles in complex and dynamic environments. Although the IQL algorithm demonstrates superior stability and convergence speed under certain conditions, further optimization is needed to enhance its adaptability and decision-making speed in highly complex, dynamic environments. Future research should focus on improving the algorithm's robustness in such environments to ensure that autonomous systems can achieve efficient and stable path planning across various real-world scenarios. This will not only increase the practical value of the algorithm but also advance the development of autonomous robotics and vehicle technologies.

\section{Conclusion}\indent

We propose an Improved Q-Learning (IQL) framework that leverages the Path Adaptive Collaborative Optimization (PACO) algorithm to optimize the initial values of the Q-table. This optimization enables the agent to quickly adapt and select optimal paths, effectively addressing key challenges in path planning within complex environments, such as slow convergence, susceptibility to local optima, and the difficulty of precise environment modeling. Furthermore, we incorporate a Utility-Controlled Heuristic (UCH) reward function mechanism to alleviate reward function sparsity, reduce exploration difficulties, and enhance reward evaluation. Compared to traditional methods, our IQL algorithm significantly improves performance, speed, and accuracy, effectively meeting the demands of path planning and obstacle avoidance. Experimental results indicate that our model demonstrates strong generalization capabilities across various scenarios.

\section*{Declarations}

\begin{itemize}
\item Funding\indent

This work was supported in part by the National Natural Science Foundation of China under Grant 52374123, in part by the Basic Scientific Research Project of the Liaoning Provincial Department of Education under Grant LJ212410147013 and Grant LJ212410147019, and in part by Liaoning Revitalization Talents Program under Grant XLYC2211085.

\item Conflict of interest/Competing interests (check journal-specific guidelines for which heading to use)\indent

The authors declare no competing financial interests.

\item Author Contributions\indent

\textbf{Ruiyang Wang}: Propose and design research, Collect and analyze data, Contribute design ideas, and Write the main parts. \textbf{Haonan Wang}: Participated in design, Provided technical support, Assisted in data analysis and Discussion. \textbf{Wei Liu} and \textbf{Guangwei Liu}: Funding acquisition, Project management, Review and Editing of writing.

\end{itemize}

\begin{figure}[h]
    \centering
    \begin{minipage}{0.3\textwidth}
        \includegraphics[width=\linewidth]{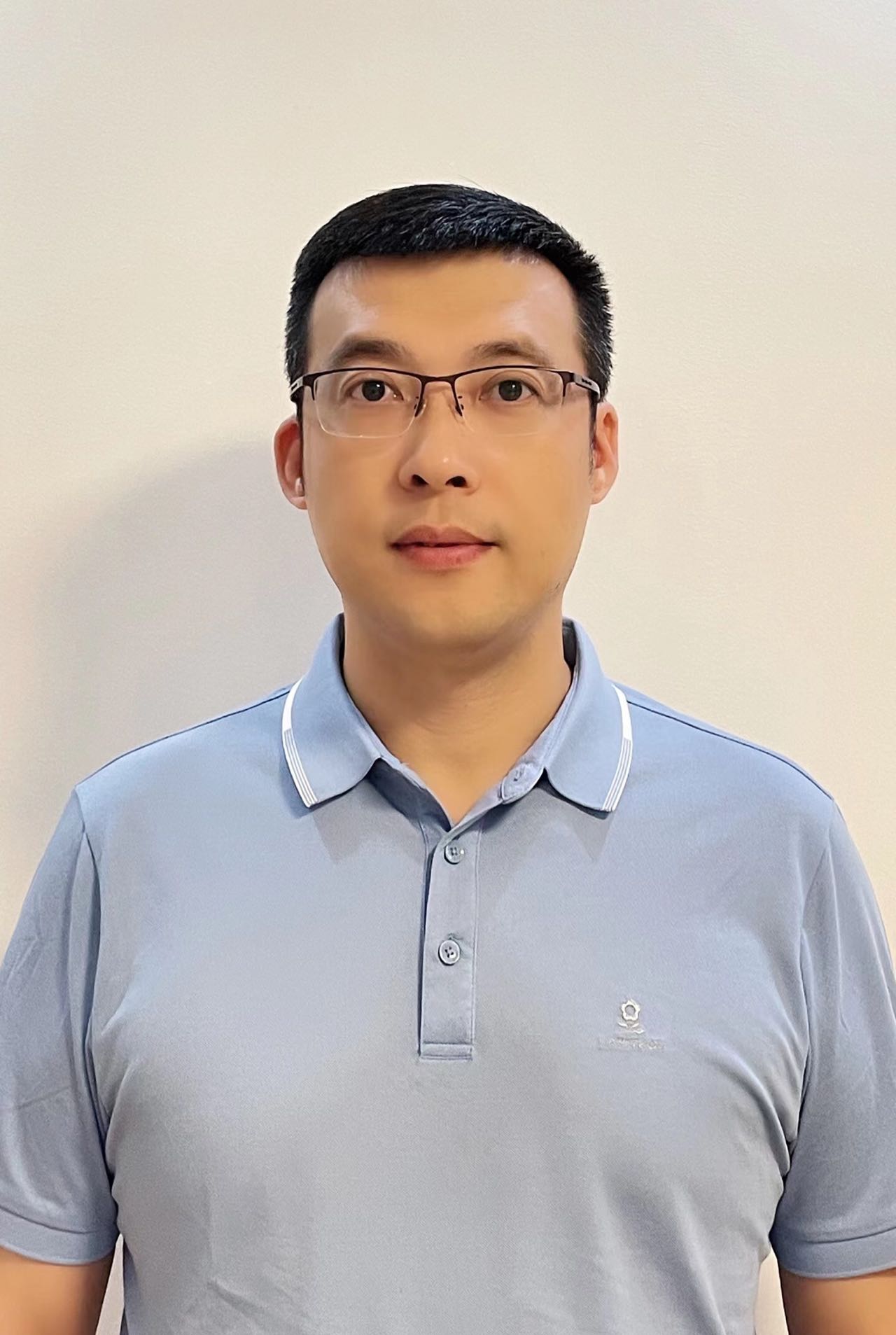} 
    \end{minipage}%
    \begin{minipage}{0.75\textwidth}
    Wei Liu received a B.S. in Mathematics Education from Shenyang Normal University (2000), an M.S. in Computer Application Technology (2008), and a Ph.D. in Mining Engineering (2019), both from Liaoning Technical University. He is currently an Associate Professor and Deputy Director at the Institute of Mathematics and System Science at Liaoning Technical University. A member of several professional societies, his research focuses on machine learning, deep neural networks, and intelligent dispatching in mining. Liu has published over 30 articles in journals like the Journal of Computer Science and has participated in numerous national and provincial projects, leading four enterprise initiatives. He has received multiple awards, including several from the China Coal Industry Science and Technology Awards.
  
    \end{minipage}
\end{figure}

\author{}
\date{}
\maketitle

\begin{figure}[h]
    \centering
    \begin{minipage}{0.3\textwidth}
        \includegraphics[width=\linewidth]{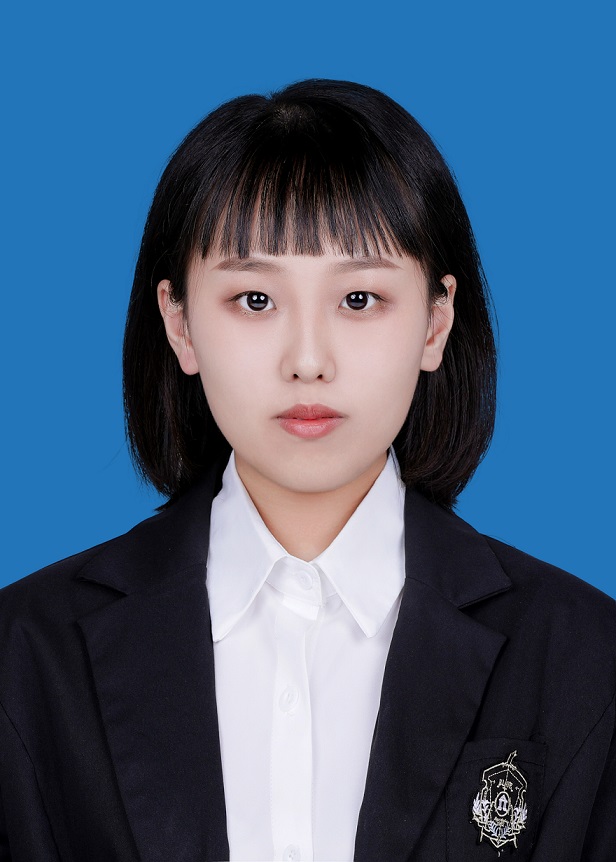} 
    \end{minipage}%
    \begin{minipage}{0.75\textwidth}
    Ruiyang Wang received a B.S. degree in information and computing science from Liaoning Technical University, Liaoning, China, in 2023. She is working toward an M.S. in mathematics major with Liaoning Technical University, Liaoning, China. Her research interests include path planning, deep learning, neural networks, and deep reinforcement learning.

    \end{minipage}
\end{figure}

\author{}
\date{}
\maketitle

\begin{figure}[h]
    \centering
    \begin{minipage}{0.3\textwidth}
        \includegraphics[width=\linewidth]{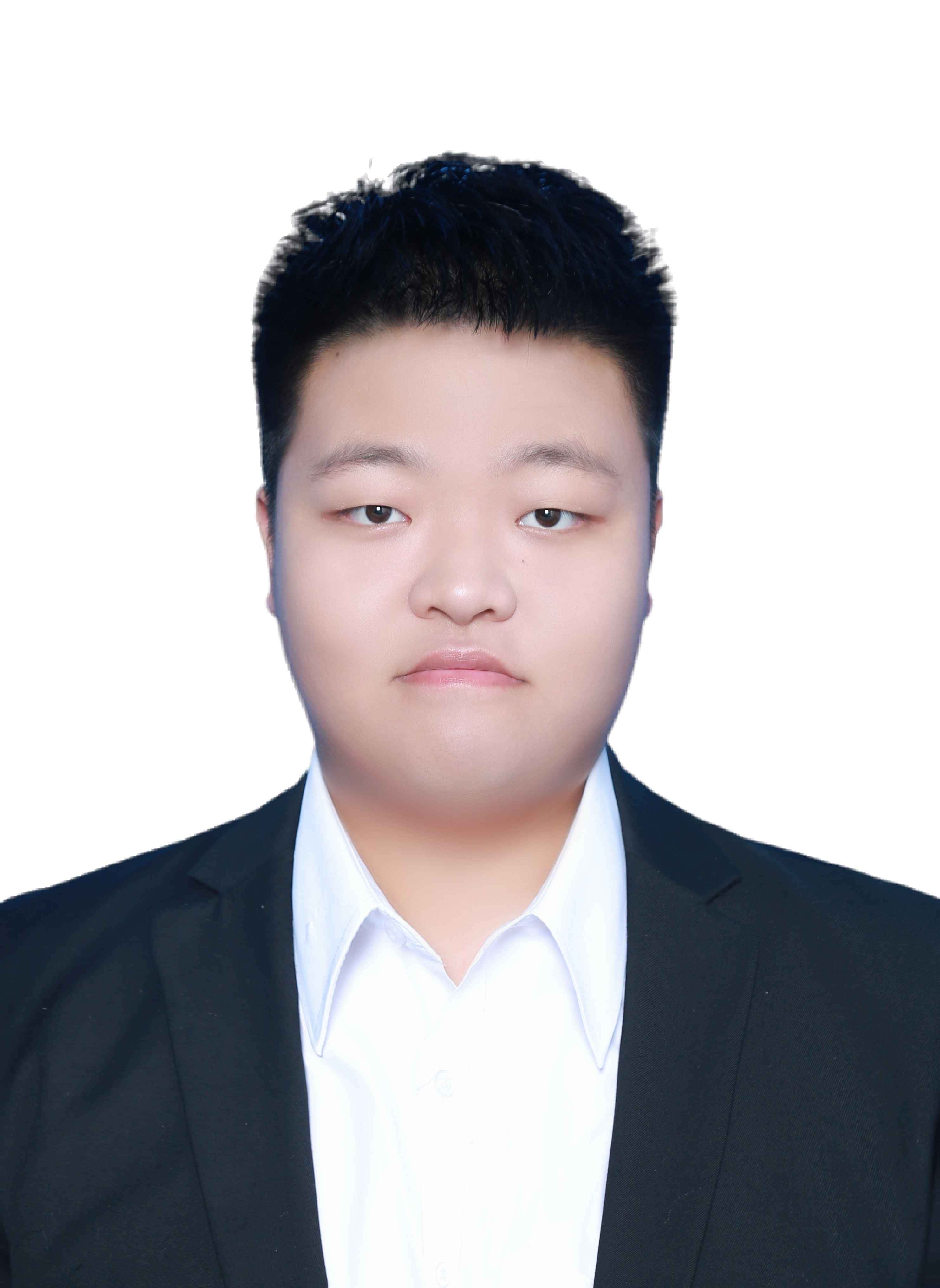} 
    \end{minipage}%
    \begin{minipage}{0.75\textwidth}
    Haonan Wang received a B.S. degree in information and computing Science from Liaoning Technical University, Fuxin, Liaoning, China. He is currently studying M.S. degree in Computer Science at Johns Hopkins University, Maryland, USA. His research interests are the intersection of Human-computer interaction and Machine Learning/ Natural Language Processing, especially Large Language Models and Agents for Human-centered evaluation in interaction systems.

    \end{minipage}
\end{figure}

\author{}
\date{}
\maketitle

\begin{figure}[h]
    \centering
    \begin{minipage}{0.3\textwidth}
        \includegraphics[width=\linewidth]{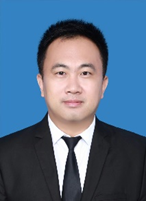} 
    \end{minipage}%
    \begin{minipage}{0.75\textwidth}
    Guangwei Liu is now a professor at the College of Mines, Liaoning Technical University, Liaoning, China. He is a part-time National Engineering Research Center for Computer Software professor. He was awarded the National Coal Youth Science and Technology Prize, and May 4th Youth Medal, and selected as one of the "One Hundred Million Talents" projects in Liaoning Province. His research interests include intelligent mining in open pit mines, system optimization, and digital mine technology. He has presided over many national and provincial key projects and won 4 first prizes and 6-second prizes of provincial and ministerial level science and technology awards. He has published more than 30 papers, 4 authorized patents, and 12 software copyrights.

    \end{minipage}
\end{figure}

\end{document}